\documentclass[letterpaper]{article} 
\makeatletter
\def\copyright@text{}
\usepackage{aaai2026}  
\usepackage{times}  
\usepackage{helvet}  
\usepackage{courier}  
\usepackage[hyphens]{url}  
\usepackage{graphicx} 
\usepackage{natbib}  
\usepackage{caption} 
\usepackage{algpseudocode}
\usepackage{makecell}
\usepackage{pgfplots}
\pgfplotsset{compat=1.18}
\usepackage{amsmath}

\usepackage[most]{tcolorbox}   
\usepackage{inconsolata}
\usepackage{xcolor}

\usepackage{amsmath}
\usepackage{amssymb}
\usepackage{amsfonts}
\usepackage{enumitem}
\usepackage{upgreek}

\tcbset{
  bigbox/.style={
    enhanced,
    breakable,
    colback=white,
    colframe=gray!50!black,
    boxrule=0.6pt,
    arc=2mm,
    left=6pt,right=6pt, top=6pt, bottom=6pt,
    coltitle=white,
    fonttitle=\bfseries,
  },
  sectiontitle/.style={
    colback=gray!10,    
    boxrule=0pt,
    arc=1mm,
    left=2pt,right=2pt,top=1pt,bottom=1pt,
    fontupper=\bfseries
  }
}
\usepackage{algorithm}

\usepackage{newfloat}
\usepackage{listings}
\DeclareCaptionStyle{ruled}{labelfont=normalfont,labelsep=colon,strut=off} 
\floatstyle{ruled}
\newfloat{listing}{tb}{lst}{}
\floatname{listing}{Listing}
\title{Toward Deployable Multi-Robot Collaboration via a Symbolically-Guided Decision Transformer}
\author {
    Rathnam Vidushika Rasanji\textsuperscript{\rm 1},
    Jin Wei-Kocsis\textsuperscript{\rm 1}\thanks{Corresponding author},
    Jiansong Zhang\textsuperscript{\rm 2},
    Dongming Gan\textsuperscript{\rm 3},
    \newline Ragu Athinarayanan\textsuperscript{\rm 3}, 
    Paul Asunda\textsuperscript{\rm 4}
}
\affiliations {
    \textsuperscript{\rm 1}School of Applied and Creative Computing, Purdue University\\
    \textsuperscript{\rm 2}School of Construction Management Technology, Purdue University\\
    \textsuperscript{\rm 3}School of Engineering Technology, Purdue University\\
    \textsuperscript{\rm 4}Department of Technology Leadership and Innovation, Purdue University\\
    \{vrathnam, kocsis0, zhan3062, dgan, rathinar, pasunda\}@purdue.edu
}

\begin{document}

\maketitle

\begin{abstract}
Reinforcement learning (RL) has demonstrated great potential in robotic operations. However, its data-intensive nature and reliance on the Markov Decision Process (MDP) assumption limit its practical deployment in real-world scenarios involving complex dynamics and long-term temporal dependencies, such as multi-robot manipulation. Decision Transformers (DTs) have emerged as a promising offline alternative by leveraging causal transformers for sequence modeling in RL tasks. However, their applications to multi-robot manipulations still remain underexplored. To address this gap, we propose a novel framework, Symbolically-Guided Decision Transformer (SGDT), which integrates a neuro-symbolic mechanism with a causal transformer to enable deployable multi-robot collaboration. In the proposed SGDT framework, a neuro-symbolic planner generates a high-level task-oriented plan composed of symbolic subgoals. Guided by these subgoals, a goal-conditioned decision transformer (GCDT) performs low-level sequential decision-making for multi-robot manipulation. This hierarchical architecture enables structured, interpretable, and generalizable decision making in complex multi-robot collaboration tasks. We evaluate the performance of SGDT across a range of task scenarios, including zero-shot and few-shot scenarios. To our knowledge, this is the first work to explore DT-based technology for multi-robot manipulation. 

\end{abstract}


\section{Introduction}

Multi-robot systems (MRSs) are increasingly deployed in high-impact domains such as construction~\cite{KalLocHar22}, manufacturing~\cite{JiLeeYoo21}, surgical procedures~\cite{howe1999robotics}, and agricultural operations~\cite{oliveira2021advances}. By enabling multiple robots to operate cooperatively in shared workspaces, MRSs can improve efficiency, robustness, and adaptability. However, safe and effective collaboration demands clearly defined objectives, efficient coordination strategies, and the ability to adapt to unforeseen scenarios. In real-world deployments, environmental and task variations frequently arise, requiring robots to plan and adapt in real time while leveraging prior experience. This interplay of control and learning remains one of the most critical and challenging capabilities in MRS design.

Reinforcement learning (RL) methods have demonstrated strong performance in robotic decision-making~\cite{gao2020deep, joshi2020robotic, deng2019deep, margolis2024rapid} but face two practical limitations in multi-robot settings: (1) high data requirements from extensive online interaction, and (2) reliance on the MDP assumption, which limits temporal reasoning over extended action sequences. Different offline RL approaches have been developed to mitigate data demands, but remain sensitive to suboptimal demonstrations and lack explicit task-level planning capabilities. Decision Transformers (DTs) have recently emerged as a promising alternative by reframing reinforcement learning as a sequence modeling problem, enabling policies to be trained from offline trajectories without explicit value function estimation~\cite{chen2021decisiontransformerreinforcementlearning}. By conditioning on return-to-go (RTG), states, and actions, DTs leverage the modeling power of causal transformers to capture long-range temporal dependencies. In~\cite{GajZurPie24}, a multi-goal DT was applied to single-robot operation tasks by incorporating goal information in the observations. In~\cite{BousVezzRao23}, a visual DT was proposed to process action-labeled visual experiences for multi-embodiment multi-task single-robot manipulation. A DT agent with natural language instructions replacing the RTG was presented in~\cite{Mansour25} for multi-task single-robot manipulation. While DTs have demonstrated encouraging results in single-robot operation tasks, their direct application to multi-robot manipulation remains largely underexplored.

To address this gap, we propose a Symbolically-Guided Decision Transformer (SGDT) framework that integrates a neuro-symbolic mechanism with a causal transformer to enable deployable multi-robot collaboration. In SGDT, a neuro-symbolic planner, which is developed by employing Planning Domain Definition Language (PDDL)~\cite{GhaHowKno98}, generates a high-level task-oriented plan composed of symbolic subgoals. Guided by these subgoals, a goal-conditioned decision transformer (GCDT) performs low-level sequential decision-making for multi-robot manipulation. This hierarchical architecture enables structured, interpretable, and generalizable decision-making in complex coordination tasks. We evaluate SGDT across diverse multi-robot manipulation scenarios, including zero-shot and few-shot settings. To our knowledge, this is the first work to explore DT-based technology for multi-robot manipulation. Our main contributions are twofold: (1) developing a novel, cost-effective, and task-oriented neuro-symbolic DT framework for multi-robot collaboration, with broad applicability across industrial domains; and (2) conducting comprehensive evaluations in different task scenarios.

The next section introduces our proposed SGDT framework. The following section shows the comprehensive performance evaluations of our proposed SGDT framework across different multi-robot manipulation task scenarios. Conclusions are presented in the last section.

\section{Proposed SGDT Framework}
\begin{figure*}[ht]
\centering
  \includegraphics[width=0.9\linewidth]{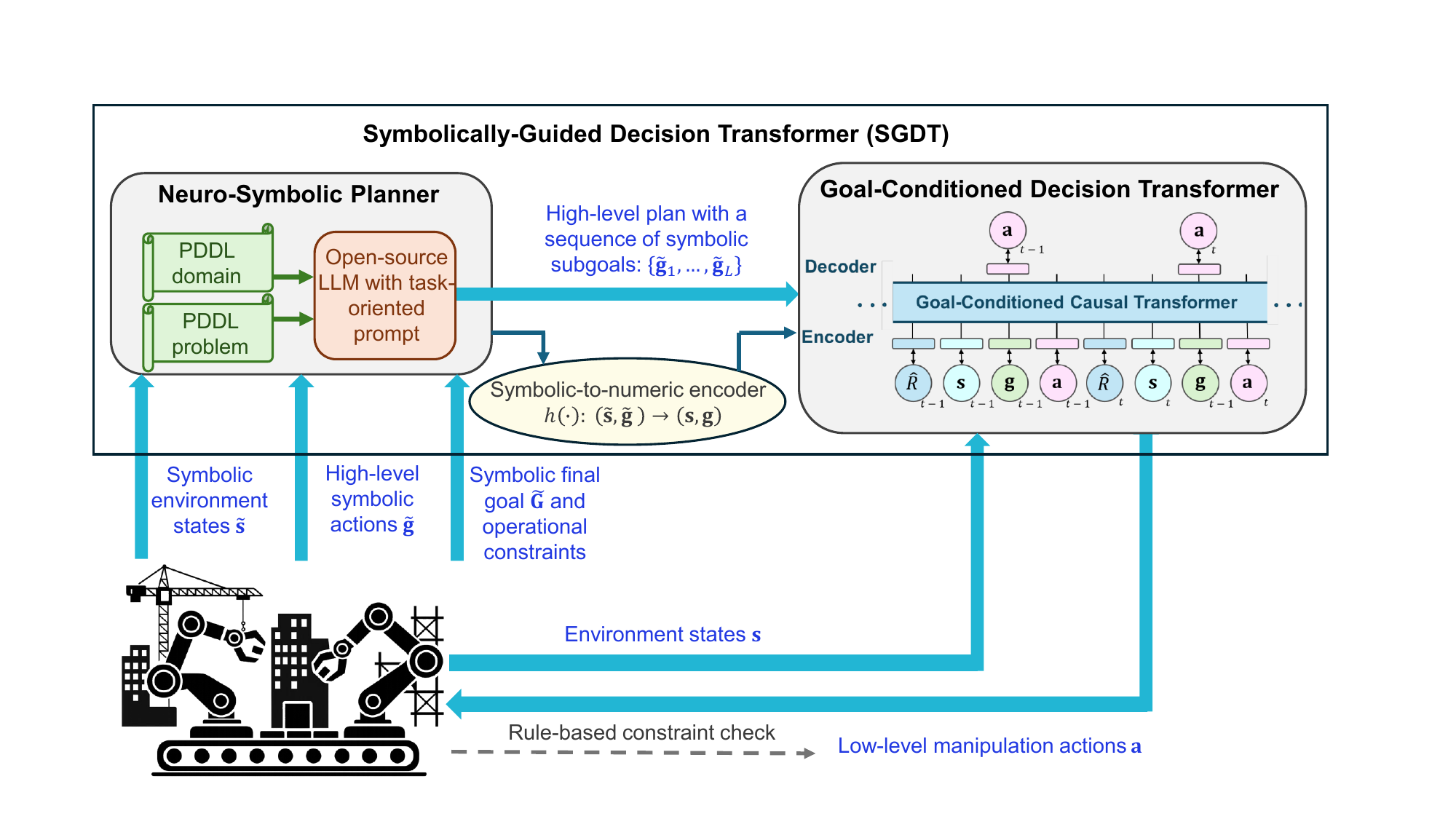}
  \caption {Overview of our proposed SGDT framework.}
  \label{fig:overview}
\end{figure*}

In our work, we focus on a cooperative multi-robot manipulation problem in which each robot agent collaborates with others to complete tasks over a finite time horizon $T$. In our current setup, the robot agents share a common observation space $\mathcal{O}$, assumed to be equivalent to the state space $\mathcal{S}$ of the collaboration environment, and collaboratively determine actions $\mathbf{a}\in\mathcal{A}$ for motion planning of their robot arms, with the objective of achieving the final goal while satisfying physical and operational constraints such as collision avoidance and inverse kinematics feasibility. To achieve this, our proposed SGDT framework enables collaborative decision making on $\mathbf{a}\in\mathcal{A}$ and its overall architecture is illustrated in Fig.~\ref{fig:overview}. As shown in Fig.~\ref{fig:overview}, SGDT framework mainly comprises a neuro symbolic planner and a goal-conditioned decision transformer (GCDT). 


\subsection{Neuro-Symbolic Planner}


In our framework, a neuro-symbolic planner performs high-level task-oriented plan reasoning. As illustrated in Fig.~\ref{fig:overview}, the multi-robot collaboration environment is first abstracted into symbolic states $\tilde{\mathbf{s}}$, representing all possible configurations of the robotic task domain using a predefined set of variable types, and symbolic high-level actions $\tilde{\mathbf{g}}$, capturing feasible subgoals of the robot arm operations in the given task. These symbolic elements are encoded via \textit{predicates} and \textit{actions} in a \textit{PDDL domain}, which also specifies the environment dynamics through operational \textit{preconditions} and \textit{effects} of each action. PDDL domain further defines \textit{parameters} that serve as placeholders for task-specific entities. When combined with a \textit{PDDL problem}, these placeholders are instantiated into concrete \textit{objects} to describe the target task scenario. The PDDL problem also specifies the \textit{initial state} and the final \textit{goal} $\tilde{\mathbf{G}}$.

Given these specifications, the planner generates an optimal plan comprising a sequence of $L$ high-level symbolic actions, \(\pi^{*} = \left\{\tilde{\mathbf{g}}_1, \cdots, \tilde{\mathbf{g}}_L\right\}\), which minimizes the symbolic plan cost \(\sum_{l=1}^{L} c\left(\tilde{\mathbf{g}}_l\right)\) when transforming the initial state into the final goal state while satisfying all preconditions, effects, and operational constraints. In our current work, we focus on minimizing the length of \(\pi^{*}\) by assigning a uniform cost \(c(\tilde{\mathbf{g}}_l) = 1\) to each symbolic high-level action. Under this formulation, the total plan cost \(\sum_{l=1}^{L} c(\tilde{\mathbf{g}}_l)\) equals the number of actions \(L\), and the planner seeks the shortest sequence of symbolic actions that achieves the goal. To address complex multi-robot manipulation plans, we employ the open-source LLM, LLaMA3, as the planner, using task-oriented prompts that incorporate the PDDL domain and PDDL problem along with task-specific context. The task-specific context not only provides additional environmental and operational details for initial plan generation, but also serves as a flexible interface for operator-provided guidance during replanning. Since this information is expressed in natural language, it enables a practical and intuitive way for human operators to remain on the loop (i.e., supervising the MRSs at a high level), providing timely intervention to the plan in response to evolving task conditions. Our primary motivation for adopting an open-source LLM is to enable cost-effective deployment and ensure reproducibility in real-world MRSs for industrial applications, without reliance on commercial LLM APIs. The LLM-based planner produces the optimal plan $\pi^{*}$, a sequence of high-level actions $\tilde{\mathbf{g}}_l$ that serve as subgoals for low-level manipulation. To interface with the GCDT, each symbolic subgoal $\tilde{\mathbf{g}}_l$ is transformed into its corresponding numerical form via a symbolic-to-numeric encoder $h(\cdot)$. This encoder maps symbolic subgoals to numerical parameters in the state space. In our current work, numerical subgoals are defined as the expected coordinates of the robot arms in the manipulation environment. The same $h(\cdot)$ is also applied to transform the symbolic state $\tilde{\mathbf{s}}$ into the numerical state $\mathbf{s}$.





\subsection{Goal-Conditioned Decision Transformer}
As shown in Fig.~\ref{fig:overview}, guided by the subgoals generated by our neuro-symbolic planner, the GCDT performs sequential decision-making for low-level manipulation actions $\mathbf{a}$. To achieve this, we formulate sequential decision-making for multi-robot manipulation actions directly within a DT framework, without relying on the conventional MDP assumption. This enables the model to explore longer-range temporal dependencies and subgoal-driven context from the neuro-symbolic planner. Within the DT framework, the key components of our multi-robot manipulation problem are formulated as follows: 
\begin{itemize}
    \item The numerical subgoal $\mathbf{g}_t \in \mathcal{G}$ is defined as the numerical representation of the corresponding symbolic subgoal $\tilde{\mathbf{g}}_l$ generated by the neuro-symbolic planner, obtained via the symbolic-to-numeric encoder $h(\cdot)$. It specifies the target configuration in the state space, specifically the expected 3D coordinates of the robots’ arms, which should be achieved before transitioning to the next subgoal. In our DT formulation, $\mathbf{g}_t$ remains constant across multiple time steps $t$ when they correspond to the same symbolic subgoal $\tilde{\mathbf{g}}_l$, until that subgoal is completed.
    \item The numerical state $\mathbf{s}_t \in \mathcal{S}$ is defined as the state vector of the manipulation environment at time $t$, which is shared by the participating robots in our current settings. This vector includes the 3D coordinates of the robots’ arms, the positions of relevant objects in the environment, and a binary flag indicating whether the current subgoal $\mathbf{g}_t$ has been completed.
    \item The numerical action $\mathbf{a}_t \in \mathcal{A}$ is defined as a vector specifying the target 3D coordinates of the robots' arms at time $t$, to which they should be moved in order to progress toward the current subgoal $\mathbf{g}_t$.
    \item The time horizon $T$ defines the total number of time steps required for multi-robot manipulation to achieve the final goal $\tilde{\mathbf{G}}$. It is dynamically determined during decision making and satisfies $T \in [nL, hL]$, where $L$ is the number of subgoals, $n$ is the default minimal number of low-level manipulation actions between two consecutive subgoals, and $h$ is the maximum allowed number of such actions.
    \item The return-to-go (RTG) $\hat{R}_t$ is defined as $\hat{R}_t = \sum_{j=t}^{T} r_j$,
which is the sum of rewards $r_j$ from the current time step $t$ to the end of the time horizon $T$.  
The reward function $r_j \in \mathbb{R}$ is given by
$r: \mathcal{S} \times \mathcal{G} \times \mathcal{A} \rightarrow \mathbb{R}$,
and evaluates the effectiveness of the low-level manipulation action $\mathbf{a}_t$ taken in state $\mathbf{s}_t$ toward achieving the current subgoal $\mathbf{g}_t$ with the minimum number of time steps.  
Specifically, the reward $r_t$ is formulated as follows:\\ 
\underline{Training Stage:}
\begin{equation*}
    r_t = \frac{\exp(-d_t)}{\sum_{j=1}^{n} \exp(-d_j)},
\end{equation*}
where $d_t = \lVert \mathbf{s}_t^r - \mathbf{g}_t \rVert_2$, $\mathbf{s}_t^r$ denotes the 3D coordinates of the participating robots' arms at time $t$, and the vector $\mathbf{g}_t$ specifies the expected coordinates of the robots' arms according to the corresponding subgoal.\\
\underline{Inference Stage:}\\
The reward is updated according to the following rules. Let $N$ be the total number of time steps taken until completion of the previous subgoal ($N=0$ if the current subgoal is the first one) and $\alpha$ is the predefined cumulative reward for progressing from one subgoal to the next.\begin{enumerate}[label=(\arabic*), leftmargin=*, align=left]
 \item If $t = N+qn+1$, where $0\leq q\leq \frac{h}{n}$,
 \[
 r_t = 
\begin{cases}
\frac{\alpha}{n}, & \text{if } q=0 \\
\frac{\alpha - \sum_{j=N+1}^{N+qn} r_j}{n}, & \text{otherwise}
\end{cases}
 \]

 \item If $N+qn+1 < t \leq N+(q+1)n$,
\begin{enumerate}[label=(\alph*)]
\item If $d_t \leq d_{t-1}$, then $r_t = r_{t-1}$.
\item If $d_t > d_{t-1}$, then $r_t = r_{t-1} - \beta$, where $\beta$ is a penalty for moving away from the current subgoal,
        and $r_{t+1}$ is updated as $r_{t+1} = \frac{\alpha - \sum_{j=N+qn+1}^{t} r_j}{N+(q+1)n - t}$.
\end{enumerate}
\end{enumerate}
\end{itemize}

As shown in Fig.~\ref{fig:overview}, our goal-conditioned decision transformer mainly comprises: 1) an encoder consisting of linear layers followed by an embedding layer; 2) a GPT-based causal transformer with a causal self-attention mask; and 3) a decoder. The encoder processes offline subgoal-conditioned trajectory samples, which are formed as $\hat{\tau}_{-k:t}=\{\hat{R}_k,\mathbf{s}_k,\mathbf{g}_k,\mathbf{a}_k, \dots, \hat{R}_t,\mathbf{s}_t, \mathbf{g}_t\}$, into token embeddings, which are fed into the GPT-based goal-conditioned causal transformer to predict the numerical action. The output is then passed to the decoder, which maps the predicted trajectory embeddings back to the original action space, producing the final manipulation actions for multi-robot collaborations. The decoded actions are subsequently validated and, if necessary, modified by an external rule-based constraint checker to ensure adherence to physical and operational constraints. 
\begin{algorithm}
\caption{The GCDT Training}
\label{GCDT_training}
\textbf{Input:} Offline trajectory dataset $\mathcal{D}$, GCDT model $GCDT(\cdot;\uptheta)$ with learnable parameters $\uptheta$, minibatch size $b$\\
\textbf{Output:} Trained parameters $\uptheta^*$
\begin{algorithmic}[1]
\State Initialize parameters $\uptheta$
\For{episode $m = 1$ to $\mathcal{M}$}
    \State Sample a minibatch $\big\{(\hat{\tau}^{(i)}_{-k:t},\,\mathbf{a}^{(i)}_t)\big\}_{i=1}^b$ from $\mathcal{D}$
    \State Compute loss: 
    \[
        \mathcal{L} \gets \frac{1}{b} \sum_{i=1}^{b} \left\|\, GCDT\!\left(\hat{\tau}^{(i)}_{-k:t};\,\uptheta\right) - \mathbf{a}^{(i)}_t \right\|_2^2
    \]
    \State Update $\uptheta \leftarrow \uptheta - \eta \nabla_{\uptheta} \mathcal{L}$
\EndFor
\State \Return $\uptheta^*$
\end{algorithmic}
\end{algorithm}

\subsubsection{Model training and Inference Pipeline}
The overall model training and inference procedures are shown in Algorithms~\ref{GCDT_training}~and~\ref{GCDT_inference}, respectively.

\section{Performance Evaluations}

\begin{figure}[h!]
    \centering
    \begin{minipage}[b]{0.47\linewidth}
        \includegraphics[width=\linewidth]{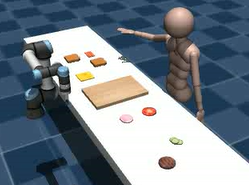}
        \caption*{(a) Sandwich Making}
    \end{minipage}
    \hfill
    \begin{minipage}[b]{0.47\linewidth}
        \includegraphics[width=\linewidth]{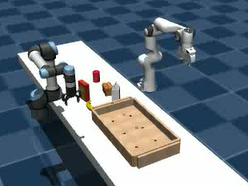}
        \caption*{(b) Grocery Packing}
    \end{minipage}
    \caption{Screenshots of tasks \emph{Sandwich Making} and \emph{Grocery Packing} in RoCoBench environment.}
    \label{fig:envs}
\end{figure}

We evaluate the performance of the proposed SGDT framework using RoCoBench, a comprehensive simulation platform designed for multi-robot collaboration tasks that span a range of coordination challenges~\cite{mandi2023rocodialecticmultirobotcollaboration}. For this study, we focus on two representative tasks, sandwich making and grocery packing, to test our framework under varying levels of complexity. Screenshots of these environments are shown in Fig.~\ref{fig:envs}. In the \emph{sandwich making} task, a humanoid robot and a UR5e robot collaborate to prepare sandwiches with recipes containing 5, 6, or 8 items. Eight food items and a cutting board are placed on a table, and robots are responsible for different items depending on their spatial arrangement. Robots must collaboratively pick up and stack items in the correct order according to the recipe. In the \emph{grocery packing} task, a Panda robot and a UR5e robot work together to place six grocery items in a box. The robots alternate turns, each starting with the closest item, requiring dynamic coordination based on relative positions. All evaluation studies are implemented on a server equipped with one NVIDIA A100 GPU, enabling efficient inference using the LLaMA-3.1 model.
\begin{algorithm}
\caption{Inference of Trained GCDT}
\label{GCDT_inference}
\textbf{Input:} Trained model $GCDT(\cdot;\uptheta^*)$, initial RTG $\hat{R}_0$, sequence of numerical subgoals $\{\mathbf{g}_{l}\}_{l=1}^{L}$, initial numerical state $\mathbf{s}_0$ \\
\textbf{Output:} Executed action sequence $\{\mathbf{a}_t\}_{t=1}^{t_{\mathrm{end}}}$
\begin{algorithmic}[1]
\State Initialize $t \gets 1$, $l \gets 1$, $\hat{R}_1 \gets \hat{R}_0$, $\mathbf{s}_1 \gets \mathbf{s}_0$, $\hat{\tau} \gets \{\mathbf{s}_1,\hat{R}_1,\mathbf{g}_{1}\}$
\While{task not terminated}
    \State Predict $\mathbf{a}_t \gets GCDT(\hat{\tau}; \uptheta^*)$
    \State Execute $\mathbf{a}_t$ and observe next state $\mathbf{s}_{t+1}$
    \State Compute reward $r_t$ and update $\hat{R}_{t+1} \gets \hat{R}_t - r_t$
    \State \textbf{Subgoal check:} \textbf{if} subgoal $\mathbf{g}_{l}$ is completed \textbf{then}
        \State \hspace{1em} \textbf{if} $l = L$ \textbf{then} $t_{\mathrm{end}} \gets t$; \Return $\{\mathbf{a}_i\}_{i=1}^{t_{\mathrm{end}}}$ \textbf{end if}
        \State \hspace{1em} \textbf{else} $l \gets l + 1$ \textbf{end if}
    \State Append $\big(\mathbf{a}_t,\hat{R}_{t+1},\mathbf{s}_{t+1}, \mathbf{g}_{l}\big)$ to $\hat{\tau}$ 
    \State $t \gets t + 1$
\EndWhile
\end{algorithmic}
\end{algorithm}
We would like to emphasize that, although our evaluation is performed on RoCoBench tasks, the collaborative manipulation patterns they capture are directly analogous to a range of industrial operations. For example, the multi-robot manipulations in \emph{sandwich making} are analogous to assembly processes in manufacturing, construction, and healthcare, while the operations in \emph{grocery packing} align with logistics and packaging workflows in warehousing, distribution, and utilities.


As shown in Fig.~\ref{fig:overview}, the proposed SGDT framework is composed of two main components: a neuro-symbolic planner and a GCDT. In this section, we evaluate the effectiveness of each component individually.

\subsection{Proposed Neuro-Symbolic Planner Evaluation}
In our proposed SGDT framework, the neuro-symbolic planner is responsible for generating high-level task-oriented plans in the form of symbolic subgoals, which guide the GCDT to determine low-level robot manipulation actions.  To illustrate how the planner operates, in the following, we present an example input and output of our neuro-symbolic planner for a sandwich-making task with a six-item sandwich recipe. The input consists of three elements: PDDL domain, which specifies object types, predicates, and actions with their preconditions and effects; PDDL problem, which defines task-specific objects, initial states, and goal conditions; and task-specific context, which provides additional operational details. Guided by this structured input, the LLaMA3-based planner generates an explicit sequence of symbolic actions. Each PICK step specifies the action, the robot, and the target object, while each PUT step specifies the action, the robot, and the relational placement between objects. This structured output provides a sequence of symbolic subgoals that define action order, navigation targets, and stacking relations. Plan generation is performed in a zero-shot manner, with up to five re-planning attempts allowed if the initial plan violates task constraints or ordering rules. In such cases, a human operator may provide additional natural-language intervention through the task-specific context. 

\begin{tcolorbox}[bigbox,title={Planner Example: Input and Output}]
{\small\tcbox[sectiontitle]{Sandwich Recipe} 
 bread\_slice1 $\rightarrow$ beef\_patty $\rightarrow$ bacon $\rightarrow$ cucumber $\rightarrow$ ham $\rightarrow$ bread\_slice2}

{\small\tcbox[sectiontitle]{PDDL Domain} 
\texttt{(define (domain sandwich-making)
    \\(:requirements :strips :typing)
    \\(:types food location robot)
    \\(:predicates       
       \\ (holding ?r - robot ?f - food)        
       \\ (on-table ?f - food)                  
       \\ (on ?f1 - food ?f2 - food)  
       \\ .....
    \\)
    \\(:action pick
        \\:parameters (?r - robot ?f - food)
        \\:precondition (and (on-table ?f) 
        \\(gripper-free ?r) (belongs-to ?f ?r))
        \\:effect (and (holding ?r ?f) 
        \\(not (on-table ?f)) (not (gripper-free ?r)))
    \\)
    \\(:action put
        \\.....
    \\)
    \\)}
\tcbox[sectiontitle]{PDDL Problem} 
\texttt{(define (problem make-bacon-sandwich)
    \\(:domain sandwich-making)
    \\(:objects 
    \\ David Chad - robot
    \\table cutting\_board - location
    \\bread\_slice1 cucumber cheese ham beef\_patty bacon tomato bread\_slice2 - food
    \\)
    \\(:init 
        \\(gripper-free David)
        \\(gripper-free Chad)\\
        \\(on-table bread\_slice1)
        \\(on-table cucumber)
        \\....
        \\(next-item bread\_slice1 cutting\_board)
        \\(next-item cucumber bread\_slice1)
          \\...
          \\(belongs-to bread\_slice1 David)
        \\(belongs-to cucumber Chad)
        \\....
        \\)
    \\(:goal 
        \\(and
            \\(on bread\_slice1 cutting\_board)
            \\(on cucumber bread\_slice1)
            \\(on cheese cucumber)
            \\......
            \\(on bread\_slice2 tomato)
        \\)
    \\)
\\)}

\tcbox[sectiontitle]{Task-Specific Context}
Generate a plan to pick an put food items according to the sandwich stack order using the above PDDL DOMAIN and PDDL PROBLEM.
Give me ONLY the sequence of actions to be performed. Make sure the actions are in 
\textit{\\PICK \textless robot\textgreater\ \textless obj\textgreater  
\\PUT \textless robot\textgreater\ \textless $obj_1$\textgreater\ \textless $obj_2$\textgreater }
\\format. DO NOT include any other comments, space lines, or symbols.

\tcbox[sectiontitle]{Generated High-Level Plan} 

\textit{PICK Dave bread\_slice1 \\
PUT Dave bread\_slice1 cutting\_board \\
PICK Chad beef\_patty \\
PUT Chad beef\_patty bread\_slice1 \\
PICK Dave bacon \\
PUT Dave bacon beef\_patty \\
PICK Chad cucumber \\
PUT Chad cucumber bacon \\
PICK Chad ham \\
PUT Chad ham cucumber \\
PICK Dave bread\_slice2 \\
PUT Dave bread\_slice2 ham
}
}
\end{tcolorbox}
We evaluate the planner using two key metrics: accuracy, defined as the percentage of generated plans that are correct and executable, and average replans, defined as the mean number of times the planner must regenerate a plan before reaching an optimal solution. These metrics capture both the reliability and robustness of the planner in collaborative multi-robot settings. Table~\ref{tab:pddl_accuracy} summarizes the results: the planner achieves a $100\%$ accuracy in sandwich-making tasks of varying complexity (5, 6, and 8 items) with no replanning required, and $90\%$ in the grocery packing task with only $0.67$ average replans. These findings demonstrate that the proposed neuro-symbolic planner is highly reliable in generating valid symbolic task plans and requires minimal human intervention, underscoring its potential for deployment in real-world industrial collaborative robotics.

 \begin{table}[h!]
\centering
\setlength{\tabcolsep}{4pt} 
\renewcommand{\arraystretch}{1.1} 
\begin{tabular}{|c|c|c|c|c|}
\hline
Tasks & \makecell{Sandwich\\making\\(5-item)} & \makecell{Sandwich\\making\\(6-item)} & \makecell{Sandwich\\making\\(8-item)} & \makecell{Grocery\\packing\\}  \\
\hline
\makecell{Accuracy\\Replans}  & \makecell{100\%\\0} & \makecell{100\%\\0} & \makecell{100\%\\0} & \makecell{90\%\\0.67} \\
\hline
\end{tabular}
\caption{Evaluation of the proposed neuro-symbolic planner.}
\label{tab:pddl_accuracy}
\end{table}

\subsection{Proposed GCDT Evaluation}
Guided by the sequence of symbolic subgoals from the neuro-symbolic planner, the GCDT operates conditioned on their corresponding numerical subgoals, represented as the expected 3D coordinates of the robot arms in the manipulation environment. These coordinates are generated through a symbolic-to-numeric encoder, which serves as an intermediate layer outside both the neuro-symbolic planner and the GCDT as shown in Fig.~\ref{fig:overview}. Based on these coordinates, the GCDT then determines the low-level manipulation actions, which are waypoint trajectories of the participating robots' arms in the considered task.


To evaluate the performance of the proposed SGDT, we first compare task success rates with and without conditioning on high-level subgoals. The comparison results for sandwich-making tasks are summarized in Table~\ref{tab:with_without_subgoal}. Task success rate is defined as the ratio of subgoals reached within a $0.1$ distance threshold to the total number of subgoals specified in the PDDL plan. These results demonstrate that conditioning on high-level subgoals substantially improves performance, particularly as task complexity grow. The subgoal-guided design alleviates the long-term temporal memory burden by providing intermediate symbolic anchors, enabling the GCDT to maintain reliability and efficiency even in extended manipulation sequences. These results demonstrate that conditioning on subgoals improves performance, especially in longer-horizon tasks, validating the GCDT design for strengthening long-term temporal memory. Additionally, for the subgoal-conditioned case, the average trajectory lengths are $54$, $66$, and $89$ time steps for the 5-item, 6-item, and 8-item sandwich tasks, respectively, indicating that the framework not only achieves higher success rates but also maintains efficient execution.

\begin{table}[h!]
\centering
\begin{tabular}{|c|c|c|c|}
\hline
Methods & 5-item & 6-item & 8-item \\ \hline
Without Subgoals & 43.08\% & 28.33\% & 29.55\% \\ \hline
With Subgoals    & 95.33\% & 92.77\% & 80.70\% \\ \hline
\end{tabular}
\caption{Task success rates of SGDT with and without conditioning on symbolic subgoals in sandwich-making tasks.}
\label{tab:with_without_subgoal}
\end{table}

\begin{figure}[h!]
\centering
\begin{tikzpicture}
\begin{axis}[
    width=8cm,
    height=5cm,
    xlabel={Data samples},
    ylabel={Task Success Rate (\%)},
    grid=major,
    ymin=50, ymax=100,
    xmin=0, xmax=100,
    xtick={0,10,...,100},
    ytick={50,60,...,100},
    thick,
    mark size=2pt,
    mark options={solid},
    legend style={at={(0.02,0.98)}, anchor=north west, fill=none, draw=none}
]

\addplot[color=blue, mark=*]
coordinates {
    (1,52.5) (5,60.25) (10,66.67) (20,77.78) (30,80.00)
    (40,82.77) (50,83.89) (60,78.88) (70,84.44)
    (80,91.10) (90,86.11) (100,92.77)
};
\addlegendentry{Grocery Packing}

\addplot[color=red, mark=square*]
coordinates {
    (1,53.33) (5,55) (10,60) (20,64.17) (30,65.83)
    (40,70) (50,64.17) (60,73.33) (70,71.67)
    (80,72) (90,74.17) (100,80)
};
\addlegendentry{Sandwich Making}

\end{axis}
\end{tikzpicture}
\caption{Task success rates vs. number of fine-tuning samples in cross-task adaptation. Blue: pretrained on the sandwich-making task, fine-tuned on the grocery-packing task. Red: pretrained on the grocery-packing task, fine-tuned on the 5-item-sandwich-making task.}
\label{fig:cross_task}
\end{figure}
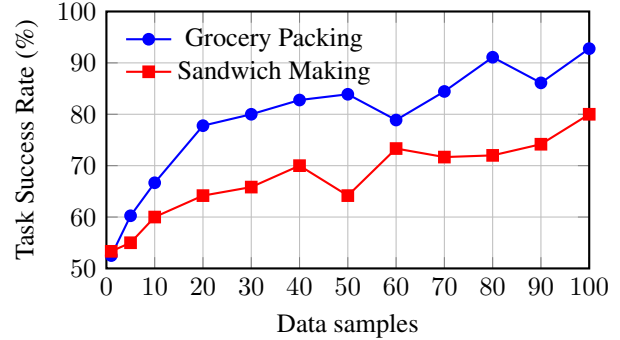

We further evaluate SGDT’s efficiency in few-shot cross-task adaptation. In the first setting, SGDT is pretrained with offline data from the sandwich-making task and then fine-tuned with a small set of grocery-packing data samples. The corresponding inference success rates are shown in the blue plot of Fig.~\ref{fig:cross_task}. We can oberve that the performance steadily improves with more fine-tuning data, exceeding $90\%$ with 100 samples, which demonstrates SGDT’s ability to adapt efficiently to a new task with limited data. In the reverse setting, SGDT is pretrained on grocery-packing data and fine-tuned for the 5-item sandwich-making task. The results, shown in the red plot of Fig.~\ref{fig:cross_task}, also reveal increasing success rates with more samples, though the gains are less steep compared to the reverse direction. This suggests that adapting from grocery packing to sandwich making is more challenging, likely due to the more complex task constraints of sandwich assembly. Together, these results validate SGDT’s few-shot adaptability across task domains, while also highlighting asymmetry in transfer efficiency depending on task complexity.

Furthermore, we evaluate SGDT’s ability in zero-shot generalization to unseen task configurations, which is a common requirement in real-world industrial settings. The model is trained using offline data from 5-item and 6-item sandwich-making tasks, and then directly evaluated on the unseen 8-item sandwich-making task without fine-tuning, achieving a success rate of $72.35\%$. This strong zero-shot performance indicates that SGDT can effectively transfer knowledge from simpler training tasks to more complex and unseen ones. To further assess its robustness, we vary the object compositions in the sandwich-making tasks by altering the number of items between training and inference. The corresponding inference success rates are summarized in Table~\ref{table:item change}. As shown in the table, SGDT achieves above $60\%$ success in most cases, demonstrating consistent robustness to unseen object sets. The only exception occurs when replacing three items in the 5-item task, where the success rate drops to $52\%$. This drop is reasonable, as the 5-item sandwich includes only three layers between the breads, making such large-scale replacements unrealistic.

\begin{table}[h!]
\centering
\setlength{\tabcolsep}{4pt} 
\renewcommand{\arraystretch}{1.1} 
\begin{tabular}{|c|c|c|}
\hline
Changed Items & \makecell{5-item Sandwich} & \makecell{6-item sandwich} \\
\hline
\makecell{1 item} & 88\% & 88.33\% \\
\hline
\makecell{2 item}   & 74\% & 65\% \\
\hline
\makecell{3 item} & 52\% & 61.67\% \\
\hline
\end{tabular}
\caption{Task Success Rate with Changed Items}
\label{table:item change}
\end{table}

\section{Conclusions}
This paper presents an initial effort to apply DT technology to multi-robot collaborations. We introduce SGDT, a novel, cost-effective, and task-oriented neuro-symbolic DT framework designed to enable deployable multi-robot collaborations, with broad applicability across industrial domains. SGDT integrates a neuro-symbolic planner with a GCDT to support structured and adaptive decision-making. We conducted comprehensive evaluations of SGDT on the RoCoBench environment, demonstrating its effectiveness across diverse multi-robot manipulation scenarios. Although the experiments were performed in simulation, the collaborative manipulation patterns are directly analogous to a wide range of industrial operations. In future work, we will evaluate and deploy SGDT in real-world industrial settings.

\bibliography{aaai2026}

\end{document}